\documentclass[10pt,twocolumn,letterpaper]{article}

\usepackage[pagenumbers]{cvpr} 

\usepackage{graphicx}
\usepackage{amsmath}
\usepackage{amssymb}
\usepackage{booktabs}
\usepackage{tabularx}
\usepackage{colortbl}
\usepackage[table]{xcolor}

\usepackage[pagebackref,breaklinks,colorlinks]{hyperref}

\usepackage[capitalize]{cleveref}
\crefname{section}{Sec.}{Secs.}
\Crefname{section}{Section}{Sections}
\Crefname{table}{Table}{Tables}
\crefname{table}{Tab.}{Tabs.}

\def\cvprPaperID{*****} 
\def\confName{CVPR}
\def\confYear{2023}

\begin{document}

\title{\LaTeX\ Author Guidelines for \confName~Proceedings}

\title{ZJU ReLER Submission for EPIC-KITCHEN Challenge 2023: \\
TREK-150 Single Object Tracking }

\author{
  Yuanyou Xu, Jiahao Li, Zongxin Yang,  Yi Yang, Yueting Zhuang\\
		 ReLER, CCAI, Zhejiang University\\
      {\tt\small
\{yoxu,xljh,yangzongxin,yangyics,yzhuang\}@zju.edu.cn}
      }
\maketitle

\begin{abstract}
    The Associating Objects with Transformers (AOT) framework has exhibited exceptional performance in a wide range of complex scenarios for video object tracking and segmentation~\cite{kristan2023tenth, yang2021associating, yangdecoupling}.
    In this study, we convert the bounding boxes to masks in reference frames with the help of the Segment Anything Model (SAM)~\cite{kirillov2023segment} and Alpha-Refine~\cite{yan2021alpha}, and then propagate the masks to the current frame, transforming the task from Video Object Tracking (VOT) to video object segmentation (VOS). 
    Furthermore, we introduce MSDeAOT, a variant of the AOT series that incorporates transformers at multiple feature scales. MSDeAOT efficiently propagates object masks from previous frames to the current frame using two feature scales of 16 and 8. 
    As a testament to the effectiveness of our design, we achieved the 1st place in the EPIC-KITCHENS TREK-150 Object Tracking Challenge.
 \end{abstract}

\section{Introduction}
\label{sec:intro}

Video object tracking is a fundamental task in computer vision that involves automatically localizing and tracking a specific object of interest across consecutive frames in a video sequence. 
It plays a crucial role in various applications, such as surveillance systems, autonomous vehicles, and video analysis. 

The tracking process typically starts with an initial bounding box or mask annotation around the target object in the first frame. Subsequently, tracking algorithms analyze subsequent frames to predict the object's location, usually by exploiting motion and appearance cues. These algorithms employ a variety of techniques, including correlation filters, deep learning-based models, particle filters, and optical flow-based methods.

In recent years, the field of video object tracking has witnessed remarkable advancements through deep learning-based approaches. SwinTrack~\cite{lin2022swintrack} stands out by leveraging transformers for feature extraction and fusion, enabling seamless interaction between the target object and the search area during tracking. Meanwhile, MixFormer~\cite{cui2022mixformer} adopts a novel approach by integrating template and test samples in the backbone network and directly regressing bounding boxes using a simple head, eliminating the need for post-processing. Another noteworthy method, SeqTrack~\cite{chen2023seqtrack}, tackles the visual tracking problem by reformulating it as a sequence generation task, enabling autoregressive prediction of the target bounding box. 

Although the methods mentioned above have yielded impressive outcomes, they are constrained in their handling of bounding boxes as both references and outputs. Real-world scenarios present challenges where bounding boxes tend to be prone to inaccuracies, potentially encompassing multiple objects. Additionally, the fixed shape of the bounding box hinders effective adaptation to target shape variations. 

Hence, tracking methods that utilize masks tend to achieve higher accuracy owing to their pixel-based alignment mechanism~\cite{li2023unified, cheng2023segment, zhou2022survey, li2022locality, yang2021collaborative}. Over the years, the Associating Objects with Transformers (AOT)~\cite{yang2021associating} series has demonstrated commendable performance in video object segmentation (VOS) tasks, prompting the natural extension of applying it to video object tracking (VOT) tasks. However, the annotation of masks poses greater difficulty compared to bounding box annotation. Fortunately, the introduction of the Segment Anything Model (SAM)~\cite{kirillov2023segment} has significantly alleviated this challenge by enabling the conversion of bounding boxes into masks with remarkable efficacy.

Specifically, we propose a novel tracking framework that leverages the AOT series to track objects in videos. We first convert the bounding boxes to masks in reference frames with the help of SAM and Alpha-Refine~\cite{yan2021alpha}, and then feed the mask and frames into the VOS model. The model then propagates the masks to the current frame and the bounding box is obtained by the mask. We also introduce MSDeAOT, a variant of the AOT series that incorporates transformers at multiple feature scales. 
Leveraging above techniques, we achieve the 1st place  in the EPIC-KITCHENS TREK-150 Object Tracking Challenge, attaining an impressive success score of 73.4\% under the multi-start evaluation protocol.

\begin{figure*}[ht]
    \vspace{-2mm}
        \centering
        \includegraphics[width=0.95\linewidth]{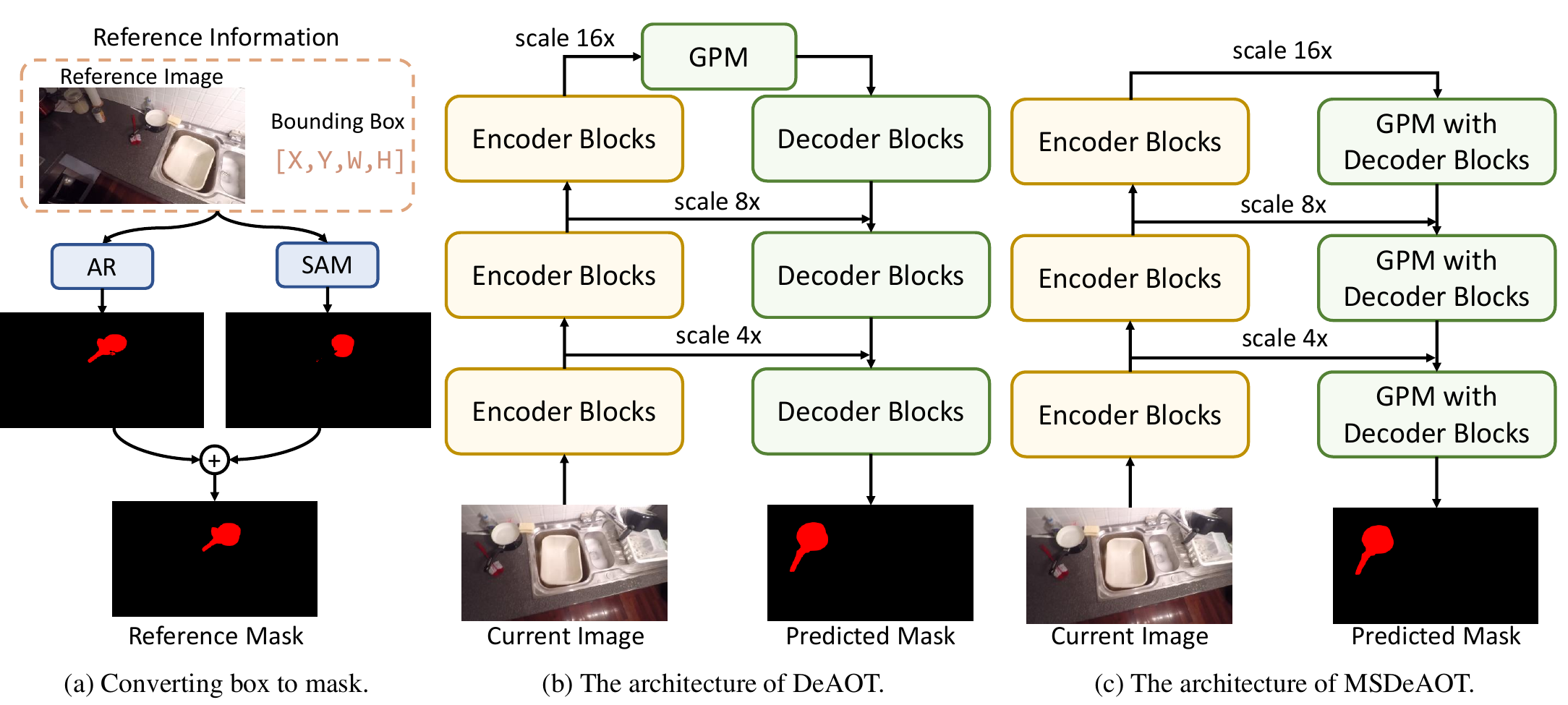}
        \caption{(a) We convert the bounding boxes to masks in reference frames with the help of SAM~\cite{kirillov2023segment} and Alpha-Refine (AR)~\cite{yan2021alpha}. (b) We feed the mask and frames into the VOS model (DeAOT)~\cite{yangdecoupling} and propagate the masks to the current frame. (c) The proposed MSDeAOT incorporates transformers at multiple feature scales.}
        \label{fig:method_overview}
    \vspace{-2mm}
\end{figure*}

\section{Method}

\subsection{Preliminaries}
\label{sec:preliminaries}

\noindent\textbf{DeAOT.} 
DeAOT is an AOT-based video object segmentation (VOS) model~\cite{yangdecoupling, yang2021associating} that incorporates an identification mechanism to associate multiple targets in a shared high-dimensional embedding space. This unique approach enables DeAOT to track multiple objects with the same efficiency as tracking a single object. To preserve object-agnostic visual information in deep propagation layers, DeAOT leverages a hierarchical Gated Propagation Module (GPM) that independently propagates both object-agnostic and object-specific embeddings from previous frames to the current frame. By utilizing GPM, DeAOT achieves effective and accurate object tracking in complex scenarios.

\noindent\textbf{Segment Anything Model (SAM).}
SAM has emerged as a prominent and influential model for image segmentation, captivating the attention of researchers in the field. Through extensive training on an extensive dataset comprising millions of images and billions of masks, SAM demonstrates exceptional proficiency. Notably, SAM excels in generating precise object masks using diverse input prompts, including points or boxes. Moreover, SAM possesses the remarkable capability to produce masks for all objects present within an image, further highlighting its versatility and effectiveness in tackling various segmentation tasks.

\noindent\textbf{Alpha-Refine (AR).}
AR represents a versatile and innovative refinement module designed to enhance visual tracking performance through precise bounding box estimation. By effectively capturing and preserving intricate spatial details, AR facilitates accurate predictions of the target's location, building upon coarse initial results. At its core, AR leverages key components such as a pixel-wise correlation layer, a corner prediction head, and an auxiliary mask head. Remarkably, AR operates as an independent module, enabling seamless integration with existing trackers in a plug-and-play manner. Notably, AR seamlessly integrates without the need for additional training or modifications to the base tracker, underscoring its adaptability and practicality.

\subsection{Multi-Scale DeAOT}
\label{sec:msdeAOT}
The whole architecture of MSAOT as shown in~\cref{fig:method_overview}\textcolor{red}{c} follows an encoder-decoder design similar to classical segmentation networks like U-Net. The encoder consists of multiple blocks that down-sample the input feature maps, yielding features at different scales. These encoder blocks provide multi-scale features that are crucial for accurate object tracking and segmentation.

In the decoder, unlike the FPN module employed in DeAOT (~\cref{fig:method_overview}\textcolor{red}{b}), the Gated Propagation Module (GPM) is integrated with multiple decoder blocks to establish the multi-scale stages of MSDeAOT. Each scale's feature maps from the encoder are fed into the corresponding stage, where the GPM module takes charge of matching the current frame with memory frames and aggregating mask information from the memory frames. The decoder blocks then decode this information.

This innovative design of multi-scale stages brings notable benefits. It effectively harnesses the potential of feature maps at different scales, in contrast to the FPN module used in DeAOT, where multi-scale feature maps solely serve as shortcut connections for residual structures. Specifically, in DeAOT, only the feature maps at the smallest scale are utilized for matching across memory frames using the GPM module. In contrast, MSDeAOT comprehensively engages feature maps from multiple scales during the matching process, thereby enhancing performance and enabling finer details of objects to be captured.
\

\section{Implementation Details}


In MSDeAOT, we employ Swin Transformer-Base~\cite{liu2021swin} as the backbones for the encoder. For the decoder, the MSDeAOT model incorporates GPM modules in multiple stages. Specifically, we set the number of layers in the GPM to 2 for the 16$\times$ scale stage and 1 for the 8$\times$ scale stage. To save computational resources, we exclude the 4$\times$ scale feature maps and instead duplicate the 16$\times$ scale feature maps twice to form the feature pyramid.

The training process comprises two phases, following the AOT framework. In the initial phase, we pre-train the model using synthetic video sequences generated from static image datasets~\cite{cheng2014global,everingham2010pascal,hariharan2011semantic,lin2014microsoft,shi2015hierarchical} by randomly applying multiple image augmentations~\cite{oh2018fast}.  In the subsequent phase, we train the model on VISOR~\cite{VISOR2022}, YouTube-VOS~\cite{xu2018youtube}, LaSOT~\cite{fan2019lasot}, MOSE~\cite{ding2023mose} and EgoTracks~\cite{tang2023egotracks},  incorporating random video augmentations~\cite{yang2020collaborative}.
For LaSOT, the mask are adopted from RTS~\cite{paul2022robust}. For EgoTracks, we sample 300 videos and generate the mask by SAM~\cite{kirillov2023segment} for each image.

During MSDeAOT training, we employ 8 Tesla V100 GPUs with a batch size of 16.  For pre-training, we use an initial learning rate of $4 \times 10^{-4}$
for 100,000 steps. For main
training, the initial learning rate is set to $2 \times 10^{-4}$, and the
training steps are 200,000.
The learning rate gradually decays to $1 \times 10^{-5}$ using a polynomial decay schedule~\cite{yang2020collaborative}.
Note that We follow the rules of the competition and use no data from the TREK150 dataset~\cite{TREK150ijcv,TREK150iccvw} for training.

During the inference process, the box references are transformed into masks, as illustrated in~\cref{fig:method_overview}\textcolor{red}{a}, and subsequently provided as input to the VOS model (MSDeAOT) along with the frames. For each frame, MSDeAOT generates predictions of the target object's mask, from which the bounding box is conveniently derived using the minimum external rectangle approach.
\section{EPIC-Kitchens Challenge: TREK-150 Single Object Tracking}
\label{sec:challenge}

\begin{table}[t]
    \centering
    \small
    \setlength{\tabcolsep}{1pt} %
      \begin{tabularx}{\linewidth}{>{\raggedright\arraybackslash}p{3cm} >{\centering\arraybackslash}X  }
   
  \hline
  Initial Mask & OPE Success Score     \\
  \hline
  AR & 61.1   \\
  SAM & 61.0  \\
  AR+SAM & 61.6  \\
  \hline

    \end{tabularx}
    \caption{Ablation study of initial mask on R50-DeAOTL~\cite{yangdecoupling}. OPE stands for one-pass evaluation.}
    \label{table:tab1_val}
\end{table}

\begin{table}[t]
    \centering
    \small
    \setlength{\tabcolsep}{1pt} %
      \begin{tabularx}{\linewidth}{>{\raggedright\arraybackslash}X >{\centering\arraybackslash}p{6cm} >{\centering\arraybackslash}X}
   
  \hline
  Backbone & Training data & OPE Success Score   \\
  \hline
  \rowcolor{gray!25} 
  R50 & VISOR~\cite{VISOR2022},MOSE~\cite{ding2023mose} & 69.4   \\
  R50 & VISOR~\cite{VISOR2022},YouTube-VOS~\cite{xu2018youtube},DAVIS~\cite{pont20172017} & 72.4   \\
  \rowcolor{gray!25} 
  R50 & VISOR~\cite{VISOR2022},YouTube-VOS~\cite{xu2018youtube},DAVIS~\cite{pont20172017},MOSE~\cite{ding2023mose} & 72.6 \\
  R50 & VISOR~\cite{VISOR2022},YouTube-VOS~\cite{xu2018youtube},DAVIS~\cite{pont20172017},MOSE~\cite{ding2023mose},EgoTracks~\cite{tang2023egotracks} & 73.4 \\
  \rowcolor{gray!25} 
  R50 & VISOR~\cite{VISOR2022},YouTube-VOS~\cite{xu2018youtube},DAVIS~\cite{pont20172017},MOSE~\cite{ding2023mose},LaSOT~\cite{fan2019lasot} & 73.3 \\
  R50 & VISOR~\cite{VISOR2022},YouTube-VOS~\cite{xu2018youtube},DAVIS~\cite{pont20172017},MOSE~\cite{ding2023mose},GOT-10k~\cite{huang2019got} & 73.3 \\
  \rowcolor{gray!25} 
  SwinB & VISOR~\cite{VISOR2022} & 70.7 \\
  SwinB & VISOR~\cite{VISOR2022},VIPSeg~\cite{miao2022large} & 71.2 \\
  \rowcolor{gray!25} 
  SwinB & VISOR~\cite{VISOR2022},MOSE~\cite{ding2023mose} & 73.3 \\

  \hline

    \end{tabularx}
    \caption{Ablation study of training data. Default model is DeAOTL and evaluated on TREK-150 test set.}
    \label{table:tab2_dataset}
\end{table}

\begin{table}[t]
    \centering
    \small
    \setlength{\tabcolsep}{1pt} %
      \begin{tabularx}{\linewidth}{>{\raggedright\arraybackslash}p{2cm} >{\centering\arraybackslash}X  >{\centering\arraybackslash}X  >{\centering\arraybackslash}X }
   
  \hline
  User & MSE Success Score  & OPE Success Score   & HOI Success Score  \\
  \hline
  JiahaoLi (Ours) & 73.4 & 75.5 & 77.1  \\
  kang & 65.3 & 67.1 & 76.2  \\
  LTMU-H-IJCV & 54.3 & 50.5 & 65.7  \\
  \hline

    \end{tabularx}
    \caption{Leaderboard on CodaLab of TREK-150 test set. MSE stands for multi-start evaluation. HOI stands for human-object interaction evaluation.}
    \label{table:final_resukts}
\end{table}

\subsection{Ablation Study}

Firstly, we conduct ablation studies on different initial masks for R50-DeAOTL~\cite{yangdecoupling}, which is trained on YouTube-VOS~\cite{xu2018youtube} and DAVIS~\cite{pont20172017}. The results, as shown in~\cref{table:tab1_val}, reveal that the initial mask generated by AR~\cite{yan2021alpha} outperforms that generated by SAM~\cite{kirillov2023segment}. Moreover, the combination of AR and SAM yields the most promising results, highlighting the significance of leveraging multiple initial masks for optimal performance.

Secondly, we perform ablation studies on various training datasets to assess their impact on the task at hand. Utilizing the DeAOTL~\cite{yangdecoupling} model with R50 and SwinB backbones, we examine the results presented in~\cref{table:tab2_dataset}, where it becomes evident that larger backbones exhibit superior performance. Notably, the inclusion of MOSE, EgoTracks, and LaSOT datasets proves to be advantageous, further enhancing the overall outcomes.

\subsection{Challenge Results}
We rank 1st place in the TREK-150 Single Object Tracking Challenge with a success score of 73.4\% under the multi-start evaluation protocol. The results are presented in~\cref{table:final_resukts}. Notably, our method achieves the highest success score in all three evaluation protocols, highlighting its robustness and versatility.

\section{Conclusion}

In this paper, we propose MSDeAOT, an AOT-based video object segmentation model. With the help of SAM and AR, we convert the bounding boxes to masks in reference frames and feed the mask and frames into the VOS model. The model then propagates the masks to the current frame and the bounding box is obtained by the mask. 
Masks offer more accurate localization than bounding boxes, and the AOT series has demonstrated remarkable performance in video object segmentation tasks. 
Our solution achieves the 1st place in the EPIC-KITCHENS TREK-150 Object Tracking Challenge with a success score of 73.4\% on the test set under the multi-start evaluation protocol.

{\small
\bibliographystyle{ieee_fullname}
\bibliography{main}
}

\end{document}